# A data-driven approach to linking design features with manufacturing process data for sustainable product development

Jiahang Li[a,*], Lucas Cazzonelli[b], Jacqueline Höllig[b], Markus Doellken[a], Sven Matthiesen[a]

[a]Karlsruhe Institute of Technology (KIT), IPEK – Institute of Product Engineering, Kaiserstr. 10, 76131 Karlsruhe, Germany
[b]FZI – Research Center for Information Technology, Haid-und-Neu-Str. 10-14, 76131 Karlsruhe, Germany
* Corresponding author. E-mail address: jiahang.li@kit.edu

**Abstract**

The growing adoption of Industrial Internet of Things (IIoT) technologies enables automated, real-time collection of manufacturing process data, unlocking new opportunities for data-driven product development. Current data-driven methods are generally applied within specific domains, such as design or manufacturing, with limited exploration of integrating design features and manufacturing process data. Since design decisions significantly affect manufacturing outcomes, such as error rates, energy consumption, and processing times, the lack of such integration restricts the potential for data-driven product design improvements. This paper presents a data-driven approach to mapping and analyzing the relationship between design features and manufacturing process data. A comprehensive system architecture is developed to ensure continuous data collection and integration. The linkage between design features and manufacturing process data serves as the basis for developing a machine learning model that enables automated design improvement suggestions. By integrating manufacturing process data with sustainability metrics, this approach opens new possibilities for sustainable product development.

## 1. Introduction

The increasing availability of process data from modern manufacturing machines, coupled with the growing adoption of Industrial Internet of Things (IIoT) technologies, is creating new opportunities for data-driven product development [1]. To effectively manage all product-related data, Product Data Management (PDM) systems provide centralized control of design and engineering data during the development process. Building on this, Product Lifecycle Management (PLM) systems extend data management across the entire product lifecycle [2]. While these systems primarily focus on the collection and management of product-related data, e.g., CAD data, the effective analysis and utilization of the collected data require advanced data-driven methods [3]. Current research on data-driven methods tends to focus on specific domains independently, such as utilizing manufacturing data for condition monitoring and predictive maintenance [4, 5]. However, the connection between design features and manufacturing process data remains insufficiently explored [3]. This gap is significant because design decisions substantially influence manufacturing outcomes such as error rates, energy consumption, and processing times [6]. A key challenge lies in systematically linking design and process data to extract actionable insights for design improvement.

### 1.1 Related work

In the manufacturing domain, real-time machine process data are primarily collected via smart sensors that meet the operational requirements of the IIoT [7, 8]. These data are transmitted using standardized protocols such as the Open Platform Communications Unified Architecture (OPC UA) standard, which ensures secure, interoperable, and real-time data exchange between machines and systems [9]. Digital twins build upon these data sources as virtual representations of physical assets, enabling advanced analysis and decision support for condition monitoring, performance optimization, and predictive maintenance of production equipment [10, 11]. Mügge et al. [12] further propose that digital twins should be used to support decision-making in the product planning phase, particularly in the context of the circular economy, but emphasize the importance of keeping the digital twin up-to-date to ensure the quality of these decisions. However, the machine process data rarely feeds back systematically into design improvements, as the focus remains on anomaly and fault detection rather than on extracting design optimization insights [13].

In the design domain, multiple methods exist for analyzing CAD data. These include feature extraction techniques such as set-based approaches and automated recognition for characterizing geometric entities, as well as neural network architectures like UV-Net, designed to process boundary representation (B-rep) data from 3D CAD models [14, 15]. Various data representations, including edge models, point clouds, and surface meshes, support comprehensive geometric analysis [16–18]. Additionally, extensive CAD datasets such as the ABC Dataset [19] or DeepCAD Dataset [20] are available for training purposes. The STEP format serves as a unified exchange standard, ensuring interoperability between different CAD programs [21]. Design knowledge is already employed to propose subsequent design steps within CAD systems [22]. Such design knowledge can be gained from experience, modeling, or testing [23–25]. If the knowledge were enriched with sustainability values that can be derived from manufacturing process data, it could directly support sustainable design practices by facilitating informed decisions during the design process.

Approaches that couple design and manufacturing, such as Design for Manufacturing (DfM) and Design for Assembly (DfA), integrate manufacturing rules into the design phase [26–28]. However, these methods primarily rely on static principles and lack continuous feedback from ongoing production [29, 30]. As a result, design decisions are often made without up-to-date production insights, missing opportunities for optimization. To address this, concurrent engineering and product production co-design approaches consider design and manufacturing simultaneously [29]. Some methods use design features to derive manufacturing steps [31], while others, such as Weisenbach et al. [32], enable remote design checks by manufacturers. Nonetheless, these approaches do not incorporate machine data analysis for integrated design feedback, and thus do not utilize data to refine design features systematically.

### 1.2 Research question

Current literature reviews show that data-driven methods in product development largely remain within their specific domains, such as design or manufacturing, while integrating these domains holds significant potential [33–35]. The problem is the lack of a systematic linkage between design features and manufacturing process data, resulting in unused potential for data-driven product design improvements. This leads to our research question:

*How can the linkage between design features and manufacturing process data be systematically established and utilized for data-driven improvement of product design?*

## 2. Methodology

To develop the approach for systematically linking design features and manufacturing process data, we employed a three-step methodology for structured data collection and connection. An overview of the data collection and connection is provided in Figure 1. Data are generated through defined activities and are captured via various implementation components within a data platform, where they are assigned to an interconnected data model. The three steps are described in detail below.

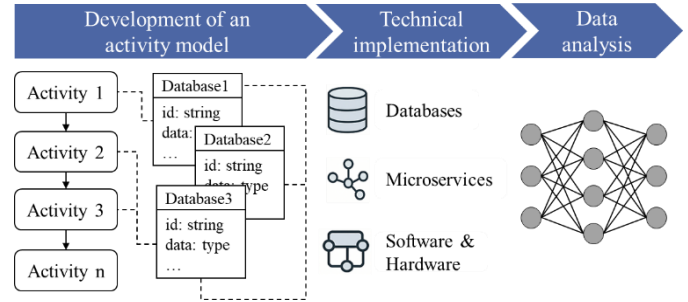

Fig. 1. Methodological framework for data collection and connection.

**Step 1: Development of an activity model for data linking**

The first step involves developing an activity model for the structured linking of design and manufacturing process data. A Unified Modeling Language (UML) activity diagram was developed during a workshop with domain experts from design, manufacturing, and IT. Various activities throughout the product lifecycle were systematically analyzed, and a unified linking schema was developed. Data linkage is achieved through the definition of common identifiers that ensure clear mapping between heterogeneous data sources.

**Step 2: Technical implementation of a data platform**

The second step involves implementing a data platform for the automated collection of various data sources (e.g., from CAD, PDM, and manufacturing equipment). Based on the schema in Figure 1, implementation components such as databases, microservices, and software and hardware modules are systematically integrated to create a unified data foundation. Databases are used to reliably store and organize the collected data. Microservices, as modular and independently deployable software units, enable smooth data exchange and specialized processing through different interfaces. Complementing these, commonly used software and hardware components are connected via Application Programming Interfaces (APIs), facilitating seamless integration with various industrial data sources from practical applications.

**Step 3: Data analysis**

A concept for an analysis process was developed to specifically evaluate the linked data using modern methods (e.g., feature extraction, machine learning) in order to identify data-driven improvement potentials and enable feedback into product design.

## 3. Result

The developed approach for systematically establishing and utilizing the linkage between design features and manufacturing process data is organized into three key phases, which are described in detail in the following subsections.

### 3.1 Activity model for data linking

An excerpt of the developed activity model for data linking from design activities and production activities was illustrated in Figure 2. Data flows from article creation through CAD design and STEP file generation to production preparation, the actual manufacturing process, and error feedback. The STEP file is used for geometric analysis since it is a unified exchange format independent of CAD systems. Machine data is collected via IIoT interfaces. Additionally, errors not detectable by machines are manually recorded by manufacturers.

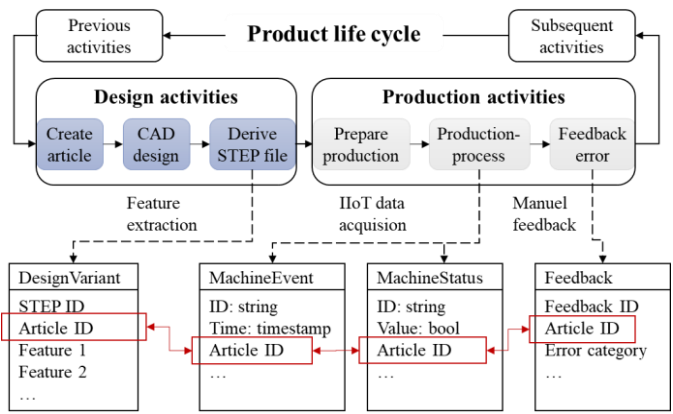

Fig. 2. Excerpt of the activity model for data linking from design and production activities.

The activities modeled generate data, which are stored in a structured form using dedicated data structures (see e.g., DesignVariant). All collected data is centrally stored on a server. The linkage is realized through common Article IDs, which are consistently used as primary or foreign keys across all data structures, such as DesignVariant, MachineEvent, MachineStatus, and Feedback, as shown in Figure 2. As a result, database queries can efficiently retrieve all relevant information related to a specific article.

### 3.2 Technical system architecture of the data platform

The technical system architecture of an integrated data platform for collecting and linking data is illustrated by a component diagram, as shown in Figure 3. The architecture is organized into three main layers: external data sources, a central server, and a dedicated machine learning server. External data originates from CAD tools, PDM systems, and machines equipped with OPC UA publishers and reverse proxies. CAD models can either be manually uploaded via a web application or automatically retrieved from the PDM system through a dedicated PDM interface. Machines produce process data, which is transmitted by an OPC UA publisher on the device and collected by an OPC UA subscriber on the server via a secure reverse proxy connection.

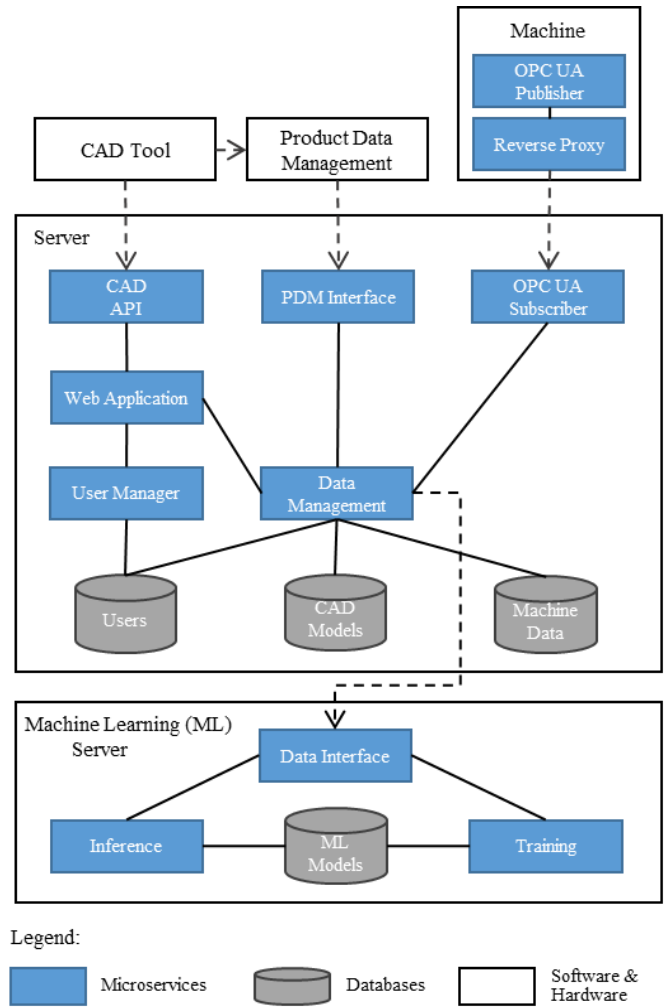

Fig. 3. Component diagram of the technical system architecture for collecting and linking design and manufacturing data.

On the server layer, various microservices, e.g., CAD API, PDM interface, and OPC UA subscriber, operate independently and manage their respective data sources. User management is handled through a centralized User Manager component, allowing for granular control of user credentials and data access. The Data Management module coordinates the ingestion, linkage, and storage of all incoming data streams, maintaining three dedicated databases for users, CAD models, and machine process data. The data management system also associates individual CAD models with corresponding manufacturing data, enabling efficient pairing for subsequent machine learning applications.

The machine learning server is connected via a standardized data interface to the platform, providing dedicated modules for model training and inference. This allows for both batch and real-time analyses, leveraging historical and current data stored within the system.

## 3.3 Machine learning based data analysis and feedback

Figure 4 illustrates the developed machine learning model for automated analysis of the linked data. The system extracts relevant features from CAD data, e.g., hole count or material thickness, from the boundary representation. These geometric and semantic features are associated, via a common article ID, with corresponding machine data, e.g., energy use, production time, or tool wear observed during manufacturing of the same part.

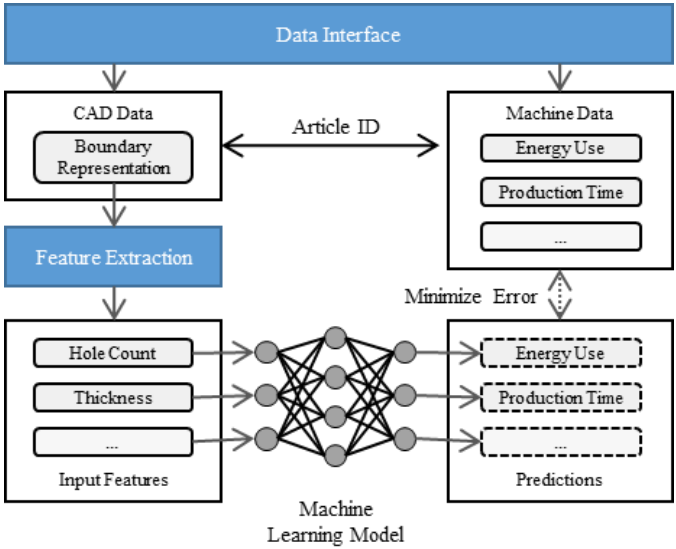

Fig. 4. Machine learning model for automated analysis of linked design features and manufacturing process data.

At the core of the system, the data interface provides the boundary representation alongside the corresponding machine data for each part. The selected features are processed and fed into the machine learning model, such as a neural network, which then generates predictions for key manufacturing parameters, e.g., energy use and production time. During the training phase, the model weights are automatically optimized by minimizing the prediction error between the forecasted and the actual observed values. In inference mode, the trained model enables data-driven feedback: its predictions can be used to estimate the likely energy consumption or production time of new designs, supporting early-stage informed decision-making.

## 4. Discussion

The research question "*How can the linkage between design features and manufacturing process data be systematically established and utilized for data-driven improvement of product design?*" was addressed through the development of a data-driven approach encompassing structured data linking, technical integration, and machine learning-based analysis, facilitating the systematic linking of design features with manufacturing process data.

While existing approaches in data-driven product development typically treat design and manufacturing separately [33–35], focusing either on CAD analysis [14–18] or operational tasks such as condition monitoring [4, 5, 8, 10–12], our work tackles the key challenge of isolated data domains. Although comparable industrial solutions that map design features to manufacturing costs using machine data exist [31], our approach advances the field by implementing a comprehensive, integrated system that ensures seamless data connectivity and generates actionable insights to support product design optimization. In contrast to traditional DfM and DfA methods, which primarily rely on static rules or unidirectional feedback [26, 27], our approach establishes a continuous, data-driven feedback loop whereby process-derived insights actively guide real-time design refinement.

With this feedback loop, decisions are no longer based on assumptions but on transparent, reproducible data analyses, allowing measurable improvements in product design and manufacturing performance. By integrating manufacturing process data with sustainability metrics, the approach opens new possibilities for design optimization and sustainable product development. For example, critical contributors to $CO_2$ emissions, such as energy consumption, can be quantified using data gathered from machine sensors during production. These data are fed into the machine learning model together with extracted design features, allowing the model to predict the sustainability impact, such as carbon footprint, of new product designs. In combination with approaches such as the one described in [22], the model can enable real-time recommendations for subsequent design steps in CAD systems to minimize environmental impact and promote more sustainable engineering decisions.

Looking ahead, the approach needs full integration into PLM infrastructures and validation through case studies using substantial real-world data. Plans include creating a dedicated database of linked design and process data in collaboration with research institutions and students, and regularly analyzing a large number of student-designed parts as part of project work. Study designs, such as [36–39], can serve as references for generating data for specific design tasks. As data volume increases, predicting manufacturing outcomes directly from design data will become more feasible, enhancing data-driven development.

A current limitation is the incomplete implementation of systematic feedback into product development, which future research should address by developing and evaluating a structured process for delivering targeted optimization recommendations to design teams.

## 5. Conclusion

The contribution of this work lies in the development of a data-driven approach for linking design features with manufacturing process data through structured data linking, technical integration, and machine learning-based analysis. By developing a connected data model and integrating diverse data sources, the approach facilitates joint analysis of design and process data, revealing critical insights into the impact of design features on key manufacturing data, including sustainability indicators. This creates a closed feedback loop that informs data-driven design improvements, shifting decisions from assumptions to evidence-based actions. While promising, further integration with PLM systems and validation with real-world data remain future tasks. The proposed approach lays a foundation for advancing data-driven product development and sustainable design optimization through deeper integration of manufacturing data.

## Acknowledgements

The authors would like to express their special thanks to TRUMPF GmbH + Co. KG, Ditzingen, Germany, particularly to Dr.-Ing. Gerhard Hammann, Mr. Jörg Heusel, Mr. Thomas Bronnhuber, and Mr. Sven King for their valuable technical support.